\documentclass[11pt]{article}
\usepackage{acl2015}
\usepackage{times}
\usepackage{url}
\usepackage{latexsym}

\usepackage{graphicx}
\usepackage{multirow}
\usepackage{amsmath,amsfonts,amssymb}

\title{Boosting Named Entity Recognition with Neural Character Embeddings}

\author{
  C\'{i}cero Nogueira dos Santos \\
  IBM Research \\
  138/146 Av. Pasteur \\
  Rio de Janeiro, RJ, Brazil \\
  {\tt cicerons@br.ibm.com} \\\And
  Victor Guimar\~{a}es \\
  Instituto de Computa\c{c}\~{a}o \\
  Universidade Federal Fluminense (UFF) \\
  Niter\'{o}i, RJ, Rio de Janeiro \\
  {\tt victorguimaraes@id.uff.br}}

\date{}

\begin{document}
\maketitle
\begin{abstract}
Most state-of-the-art named entity recognition (NER) systems rely on handcrafted features and on the output of other NLP tasks such as part-of-speech (POS) tagging and text chunking.
In this work we propose a language-independent NER system that uses automatically learned features only.
Our approach is based on the CharWNN deep neural network,
which uses word-level and character-level representations (embeddings) to perform sequential classification.
We perform an extensive number of experiments using two annotated corpora in two different languages:
HAREM I corpus, 
which contains texts in Portuguese; 
and the SPA CoNLL-2002 corpus, 
which contains texts in Spanish. 
Our experimental results shade light on the contribution of neural character embeddings for NER.
Moreover,
we demonstrate that the same neural network which has been successfully applied to POS tagging can also achieve state-of-the-art results for language-independet NER, 
using the same hyperparameters, 
and without any handcrafted features.
For the HAREM I corpus, 
CharWNN
outperforms the state-of-the-art system by 7.9 points in the F1-score for the total scenario (ten NE classes), 
and by 7.2 points in the F1 for the selective scenario
(five NE classes).
\end{abstract}


\section{Introduction} \label{sec:introduction}
Named entity recognition is a natural language processing (NLP) task that consists of finding names in a text and classifying them among several predefined categories of interest such as
person, 
organization, 
location and 
time.
Although machine learning based systems have been the predominant approach to achieve state-of-the-art results for NER,
most of these NER systems rely on the use of costly handcrafted features and on the output of other NLP tasks \cite{tjong2002conll,tjong2003conll,doddington:2004,finkel:2005,milidiu2007}.
On the other hand,
some recent work on NER have used deep learning strategies which minimize the need of these costly features \cite{chen:2010,collobert:jmlr2011,passos:2014,tang:2014}.
However,
as far as we know,
there are still no work on deep learning approaches for NER that use character-level embeddings.

In this paper we approach language-independent NER using CharWNN,
a recently proposed deep neural network (DNN) architecture that jointly uses word-level and character-level embeddings to perform sequential classification \cite{santos2014}.
CharWNN employs a convolutional layer that allows effective character-level feature extraction from words of any size.
This approach has proven to be very effective for language-independent POS tagging \cite{santos2014}.

We perform an extensive number of experiments using two annotated corpora:
HAREM I corpus,
which contains texts in Portuguese;
and the SPA CoNLL-2002,
which contains texts in Spanish.
In our experiments,
we compare the performance of the joint and individual use of character-level and word-level embeddings.
We provide information on the impact of unsupervised pre-training of word embeddings in the performance of our proposed NER approach.
Our experimental results evidence that CharWNN is effective and robust  for Portuguese and Spanish NER.
Using the same CharWNN configuration used by dos Santos and Zadrozny (2014) for POS Tagging,
we achieve state-of-the-art results for both corpora.
For the HAREM I corpus,
CharWNN outperforms the state-of-the-art system by 7.9 points in the F1-score for the \emph{total scenario} (ten NE classes),
and by 7.2 points in the F1 for the \emph{selective scenario} (five NE classes).
This is a remarkable result for a NER system that uses only automatically learned features.

This work is organized as follows.
In Section \ref{sec:model},
we briefly describe the CharWNN architecture.
Section \ref{sec:experimental_setup} details our experimental setup and Section \ref{sec:experimental_results} discuss our experimental results.
Section \ref{sec:conclusions} presents our final remarks.
\section{CharWNN} \label{sec:model}

\begin{figure*}[ht!]
\centering
\includegraphics[width=.9\textwidth]{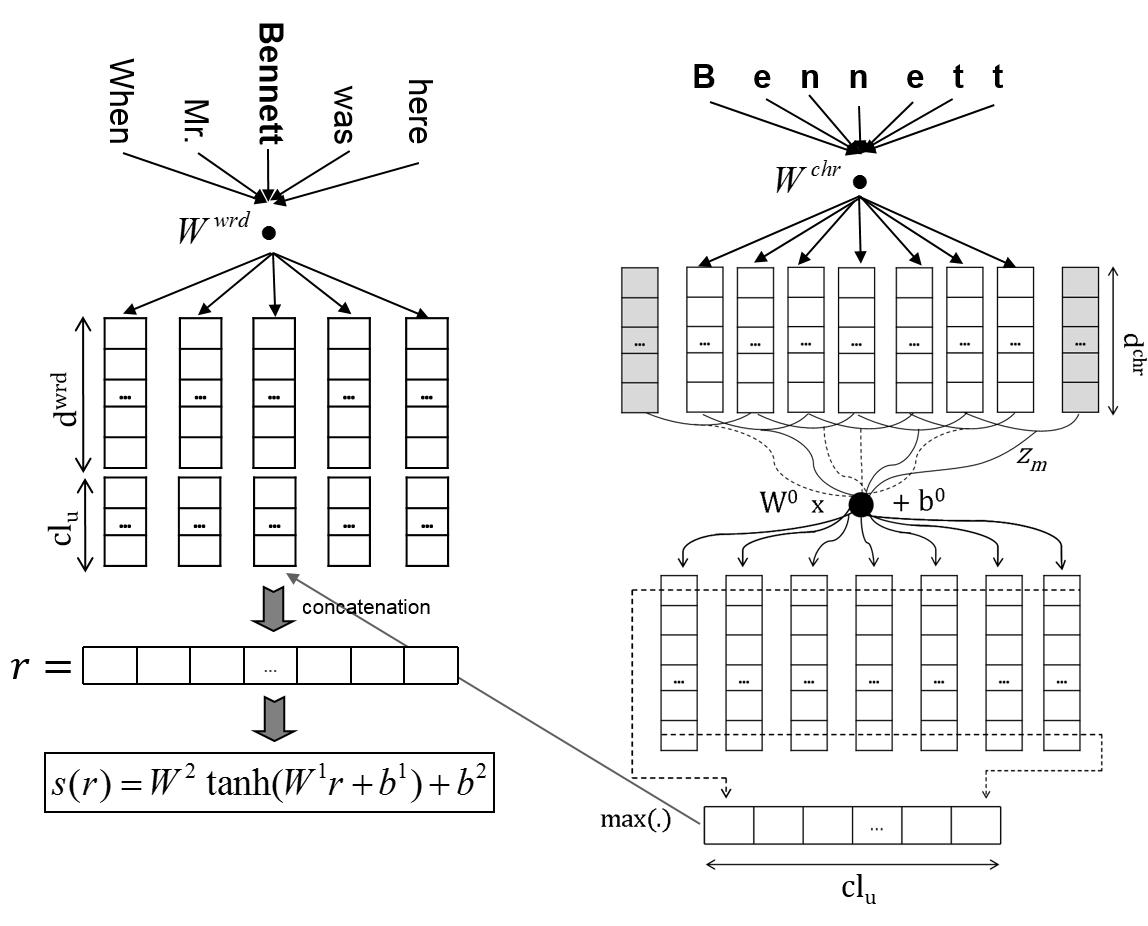}
\caption{CharWNN Architecture}
\label{fig:charwnn}
\end{figure*}

CharWNN extends Collobert et al.'s (2011) neural network architecture for sequential classification by adding a convolutional layer to extract character-level representations \cite{santos2014}.
Given a sentence,
the network gives for each word a score for each class (tag) $\tau \in T$.
As depicted in Figure \ref{fig:charwnn},
in order to score a word,
the network takes as input a fixed-sized window of words centralized in the target word. 
The input is passed through a sequence of layers where features with increasing levels of complexity are extracted.
The output for the whole sentence is then processed using the Viterbi algorithm \cite{viterbi:1967} to perform structured prediction.
For a detailed description of the CharWNN neural network we refer the reader to \cite{santos2014}.

\subsection{Word- and Character-level Embeddings}
As illustrated in Figure \ref{fig:charwnn},
the first layer of the network transforms words into real-valued feature vectors (embeddings). 
These embeddings are meant to capture morphological, 
syntactic and semantic information about the words.
We use a fixed-sized word vocabulary $V^{wrd}$,
and we consider that words are composed of characters from a fixed-sized character vocabulary $V^{chr}$.
Given a sentence consisting of $N$ words $\{w_{1}, w_{2}, ..., w_{N}\}$, 
every word $w_{n}$ is converted into a vector $u_n=[r^{wrd};r^{wch}]$,
which is composed of two sub-vectors: 
the \emph{word-level embedding} $r^{wrd}\in\mathbb{R}^{d^{wrd}}$ and the \emph{character-level embedding} $r^{wch}\in\mathbb{R}^{cl_{u}}$ of $w_n$.
While word-level embeddings capture syntactic and semantic information,
character-level embeddings capture morphological and shape information.

\emph{Word-level embeddings} are encoded by column vectors in an embedding matrix $W^{wrd}\in\mathbb{R}^{d^{wrd}\times|V^{wrd}|}$,
and retrieving the embedding of a particular word consists in a simple matrix-vector multiplication.
The matrix $W^{wrd}$ is a parameter to be learned,
and the size of the word-level embedding $d^{wrd}$ is a hyperparameter to be set by the user.

The \emph{character-level embedding} of each word is computed using a convolutional layer \cite{waibel:1989,lecun:ieee98}.
In Figure \ref{fig:charwnn},
we illustrate the construction of the character-level embedding for the word \emph{Bennett}, but the same process is used to construct the character-level embedding of each word in the input.
The convolutional layer first produces local features around each character 
of the word, 
and then combines them using a max operation to
create a fixed-sized character-level embedding of the word.

Given a word $w$ composed of $M$ characters $\{c_1, c_2, ..., c_M\}$,
we first transform each character $c_m$ into a character embedding $r^{chr}_m$.
Character embeddings are encoded by column vectors in the embedding matrix $W^{chr}\in\mathbb{R}^{d^{chr}\times|V^{chr}|}$.
Given a character $c$, 
its embedding $r^{chr}$ is obtained by the matrix-vector product:
$r^{chr} = W^{chr}v^{c}$,
where $v^{c}$ is a vector of size $\left|V^{chr}\right|$ which has value $1$ at index $c$ and zero in all other positions.
The input for the convolutional layer is the sequence of character embeddings $\{r^{chr}_1, r^{chr}_2, ..., r^{chr}_M\}$.

The convolutional layer applies a matrix-vector operation to each window of size $k^{chr}$ of successive windows in the sequence $\{r^{chr}_1, r^{chr}_2, ..., r^{chr}_M\}$.
Let us define the vector $z_m\in\mathbb{R}^{d^{chr}k^{chr}}$ as the concatenation of the character embedding $m$,
its $(k^{chr}-1)/2$ left neighbors,
and its $(k^{chr}-1)/2$ right neighbors:
\begin{displaymath}
z_m= \left(r^{chr}_{m-(k^{chr}-1)/2}, ..., r^{chr}_{m+(k^{chr}-1)/2}\right)^T
\end{displaymath}
The convolutional layer computes the $j$-th element of the vector $r^{wch}$,
which is the character-level embedding of $w$,
as follows:
\begin{equation} \label{conv_output}
[r^{wch}]_j = \max_{1 < m < M}\left[ W^{0}z_m + b^{0}\right]_j
\end{equation}where $W^{0} \in \mathbb{R}^{cl_{u} \times d^{chr}k^{chr}}$ is the weight matrix of the convolutional layer.
The same matrix is used to extract local features around each character window of the given word.
Using the max over all character windows of the word,
we extract a fixed-sized feature vector for the word.

Matrices $W^{chr}$ and $W^{0}$,
and vector $b^{0}$ are parameters to be learned.
The size of the character vector $d^{chr}$,
the number of convolutional units $cl_{u}$ 
(which corresponds to the size of the character-level embedding of a word),
and the size of the character context window $k^{chr}$
are hyperparameters.

\subsection{Scoring and Structured Inference}
We follow Collobert et al.'s \cite{collobert:jmlr2011} window approach to score all tags $T$ for each word in a sentence.
This approach follows the assumption that in sequential classification the tag of a word depends mainly on its neighboring words.
Given a sentence with $N$ words $\{w_1, w_2, ..., w_N\}$,
which have been converted to joint word-level and character-level embedding $\{u_1, u_2, ..., u_N\}$,
to compute tag scores for the $n$-th word $w_n$ in the sentence,
we first create a vector $r$ resulting from the concatenation of a sequence of $k^{wrd}$ embeddings,
centralized in the $n$-th word:
\begin{displaymath}
r=\left(u_{n-(k^{wrd}-1)/2}, ..., u_{n+(k^{wrd}-1)/2}\right)^T
\end{displaymath}
We use a special \emph{padding token} for the words with indices outside of the sentence boundaries.

Next,
the vector $r$ is processed by two usual neural network layers,
which extract one more level of representation and compute the scores:
\begin{equation} \label{word_score}
\mathbf{\mathit{s}}(w_n)=W^2h(W^1r + b^1) + b^2
\end{equation} where matrices $W^{1}\in\mathbb{R}^{hl_{u} \times k^{wrd}(d^{wrd}+cl_{u})}$ and $W^{2}\in\mathbb{R}^{\left|T\right| \times hl_{u}}$,
and vectors $b^{1}\in\mathbb{R}^{hl_{u}}$ and $b^{2}\in\mathbb{R}^{\left|T\right|}$ are parameters to be learned.
The transfer function $h(.)$ is the hyperbolic tangent.
The size of the context window $k^{wrd}$ and the number of hidden units $hl_{u}$ are hyperparameters to be chosen by the user.

Like in \cite{collobert:jmlr2011},
CharWNN uses a prediction scheme that takes into account the sentence structure.
The method uses a transition score $A_{tu}$ for jumping from tag $t \in T$ to $u \in T$ in successive words,
and a score $A_{0t}$ for starting from the $t$-th tag.
Given the sentence $[w]_1^N = \{w_1, w_2, ..., w_N\}$, 
the score for tag path $[t]_1^N=\{t_1,  t_2, ..., t_N\}$ is computed as follows:
\begin{equation} \label{sentence_score}
S\left([w]_1^N, [t]_1^N, \theta \right) =  \sum_{n=1}^{N}{\left( A_{t_{n_-1}t_n} + s(w_n)_{t_n} \right)}
\end{equation}
where $s(w_n)_{t_n}$ is the score given for tag $t_n$ at word $w_n$ and
$\theta$ is the set of all trainable network parameters $\left(W^{wrd}, W^{chr}, W^0, b^0, W^1, b^1, W^2, b^2, A\right)$.
After scoring each word in the sentence,
the predicted sequence is inferred with the Viterbi algorithm.


\subsection{Network Training}
We train CharWNN by minimizing a negative likelihood over the training set $D$.
In the same way as in \cite{collobert:jmlr2011},
we interpret the sentence score (\ref{sentence_score}) as a conditional probability over a path.
For this purpose,
we exponentiate the score (\ref{sentence_score}) and normalize it with respect to all possible paths.
Taking the log,
we arrive at the following conditional log-probability:
\begin{equation} \label{log_prob}
\begin{array}{r}
\mathrm{log} \ p\left([t]_1^N|[w]_1^N, \theta \right) =  S\left([w]_1^N, [t]_1^N, \theta \right) \\
     - \displaystyle \mathrm{log} \left( \sum_{\forall [u]_1^N \in T^{N}}{e^{S\left([w]_1^N, [u]_1^N, \theta \right)}} \right) 
\end{array}
\end{equation}

The log-likelihood in Equation \ref{log_prob} can be computed efficiently using dynamic programming \cite{collobert:aistats2011}.
We use stochastic gradient descent (SGD) to minimize the
negative log-likelihood with respect to $\theta$.
We use the backpropagation algorithm to compute the gradients of the neural network.
We implemented CharWNN using the \emph{Theano} library \cite{bergstra:scipy2010}.




\section{Experimental Setup}\label{sec:experimental_setup}
\subsection{Unsupervised Learning of Word Embeddings} \label{unsup:word:embeddings}
The word embeddings used in our experiments are initialized by means of unsupervised pre-training.
We perform pre-training of word-level embeddings using the skip-gram NN architecture~\cite{mikolov:icrl2013} available in the word2vec~\footnote{http://code.google.com/p/word2vec/} tool.

In our experiments on Portuguese NER,
we use the word-level embeddings previously trained by \cite{santos2014}.
They have used a corpus composed of the Portuguese Wikipedia,
the CE\-TENFolha\footnote{http://www.linguateca.pt/cetenfolha/} corpus
and the CETEMPublico\footnote{http://www.linguateca.pt/cetempublico/} corpus.

In our experiments on Spanish NER,
we use the Spanish Wikipedia.
We process the Spanish Wikipedia corpus using the same steps used by \cite{santos2014}:
(1) remove paragraphs that are not in Spanish;
(2) substitute non-roman characters by a special character;
(3) tokenize the text using a tokenizer that we have implemented;
(4) remove sentences that are less than 20 characters long (including white spaces) or have less than 5 tokens;
(5) lowercase all words and substitute each numerical digit by a 0.
The resulting corpus contains around 450 million tokens.

Following \cite{santos2014},
we do not perform unsupervised learning of character-level embeddings.
The character-level embeddings are initialized by randomly sampling each value from an uniform distribution: $\mathcal{U}\left(-r,r\right)$,
where $r=\sqrt{\dfrac{6}{|V^{chr}|+d^{chr}}}$.


\subsection{Corpora}

\begin{table*}[ht!]
\center
\caption{Named Entity Recognition Corpora. \label{tab:ner:corpora}}
\setlength\tabcolsep{4pt}
\begin{tabular}{llrrrr}
  \hline
   & & \multicolumn{2}{c}{\textbf{Training Data}} & \multicolumn{2}{c}{\textbf{Test Data}} \\
  \raisebox{1.5ex}{\textbf{Corpus}} & \raisebox{1.5ex}{\textbf{Language}} & \textbf{Sentenc.} &  \textbf{Tokens} & \textbf{Sentenc.} & \textbf{Tokens} \\
  \hline
  HAREM I         & Portuguese    &  4,749 &   93,125 & 3,393 & 62,914 \\
  SPA CoNLL-2002  & Spanish       &  8,323 & 264,715  & 1,517 & 51,533 \\
  \hline
\end{tabular}
\end{table*}

We use the corpus from the first HAREM evaluation \cite{santosdiana07} in our experiments on Portuguese NER.
This corpus is annotated with ten named entity categories:
Person (PESSOA), 
Organization (ORGANIZACAO),
Location (LOCAL),
Value (VALOR),
Date (TEMPO),
Abstraction (ABSTRACCAO),
Title (OBRA),
Event (ACONTECIMENTO),
Thing (COISA)
and Other (OUTRO).
The HAREM corpus is already divided into two subsets: 
First HAREM and MiniHAREM.
Each subset corresponds to a different Portuguese NER contest.
In our experiments, 
we call HAREM I the setup where we use the First HAREM corpus as the training set and the MiniHAREM corpus as the test set.
This is the same setup used by dos Santos and Milidi\'{u} (2012).
Additionally,
we tokenize the HAREM corpus and create a development set that comprises 5\% of the training set.
Table \ref{tab:ner:corpora} present some details of this dataset.

In our experiments on Spanish NER we use the SPA CoNLL-2002 Corpus,
which was developed for the CoNLL-2002 shared task \cite{tjong2002conll}.
It is annotated with four named entity categories:
Person,
Organization,
Location
and Miscellaneous.
The SPA CoNLL-2002 corpus is already divided into training, 
development and test sets.
The development set has characteristics similar to the test corpora.

We treat NER as a sequential classification problem. 
Hence, in both corpora we use the \texttt{IOB2} tagging style where:
\texttt{O},
means that the word is not a NE;
\texttt{B-X} is used for the leftmost word of a NE type \texttt{X};
and
\texttt{I-X} means that the word is inside of a NE type \texttt{X}.
The \texttt{IOB2} tagging style is illustrated in the following example.
\texttt{
\begin{small}
\begin{center}
Wolff$/$B-PER \ ,$/$O \ currently$/$O \ a$/$O \ journalist$/$O \ in$/$O Argentina$/$B-LOC \  ,$/$O 
played$/$O \ with$/$O \ Del$/$B-PER Bosque$/$I-PER \ in$/$O \ the$/$O \ final$/$O \ years$/$O \ of$/$O 
the$/$O \ seventies$/$O \ in$/$O \ Real$/$B-ORG \ Madrid$/$I-ORG
\end{center}
\end{small}
}

\subsection{Model Setup}
\label{model:setup}
In most of our experiments,
we use the same hyperparameters used by dos Santos and Zadrozny (2014) for part-of-speech tagging.
The only exception is the learning rate for SPA CoNLL-2002,
which we set to 0.005 in order to avoid divergence.
The hyperparameter values are presented in Table \ref{tab:hyperparameters}.
We use the development sets to determine the number of training epochs,
which is six for HAREM and sixteen for SPA CoNLL-2002.

\begin{table*}[ht]
\begin{center}
\caption{Neural Network Hyperparameters.}
\label{tab:hyperparameters}
\begin{tabular}{llrrr}
\hline
\bf Parameter & \bf Parameter Name & \bf CharWNN & \bf WNN & \bf CharNN\\
\hline
$d^{wrd}$ & Word embedding dimensions   & 100    & 100    & -     \\
$k^{wrd}$ & Word context window size    & 5      & 5      & 5     \\ 
$d^{chr}$ & Char. embedding dimensions  & 10     & -      & 50    \\ 
$k^{chr}$ & Char. context window size   & 5      & -      & 5     \\ 
$cl_{u}$  & Convolutional units         & 50     & -      & 200    \\ 
$hl_{u}$  & Hidden units                & 300    & 300    & 300    \\ 
$\lambda$ & Learning rate               & 0.0075 & 0.0075 & 0.0075 \\
\hline
\end{tabular}
\end{center}
\end{table*}

We compare CharWNN with two similar neural network architectures: CharNN and WNN.
CharNN is equivalent to CharWNN without word embeddings,
i.e.,  
it uses character-level embeddings only.
WNN is equivalent to CharWNN without character-level embeddings,
i.e.,  
it uses word embeddings only.
Additionally,
in the same way as in \cite{collobert:jmlr2011},
we check the impact of adding to WNN two handcrafted features that contain character-level information, 
namely capitalization and suffix.
The capitalization feature has five possible values: 
all lowercased,
first uppercased,
all uppercased,
contains an uppercased letter,
and all other cases.
We use suffix of size three. 
In our experiments,
both capitalization and suffix embeddings have dimension five.
The hyperparameters values for these two NNs are shown in Table \ref{tab:hyperparameters}.

\section{Experimental Results}\label{sec:experimental_results}

\subsection{Results for Spanish NER}
In Table \ref{tab:esp:nns},
we report the performance of different NNs for the SPA CoNLL-2002 corpus.
All results for this corpus were computed using the CoNLL-2002 evaluation script\footnote{http://www.cnts.ua.ac.be/conll2002/ner/bin/conlleval.txt}.
CharWNN achieves the best precision, 
recall 
and F1 
in both development and test sets.
For the test set,
the F1 of CharWNN is 3 points larger than the F1 of the WNN that uses two additional handcrafted features:
suffixes and capitalization.
This result suggests that,
for the NER task,
the character-level embeddings are as or more effective as the two character-level features used in WNN. 
Similar results were obtained by dos Santos and Zadrozny (2014) in the POS tagging task.

In the two last lines of Table \ref{tab:esp:nns} we can see the results of using word embeddings and character-level embeddings separately.
Both,
WNN that uses word embeddings only
and CharNN,
do not achieve results competitive with the results of the networks that jointly use word-level and character-level information.
This is not surprising,
since it is already known in the NLP community that jointly using word-level and character-level features is important to perform named entity recognition.

\begin{table*}[ht!]
\begin{center}
\caption{Comparison of different NNs for the SPA CoNLL-2002 corpus.}
\label{tab:esp:nns}
\begin{tabular}{c|c|rrr|rrr}
\hline
\multirow{2}{*}{\bf NN} & \multirow{2}{*}{\bf Features} & \multicolumn{3}{c|}{\textbf{Dev. Set}} & \multicolumn{3}{c}{\textbf{Test Set}} \\
\cline{3-8}
       & & \bf Prec. & \bf Rec. & \bf F1 & \bf Prec. & \bf Rec. & \bf F1\\
\hline
CharWNN & word emb., char emb.      & \bf 80.13 & \bf 78.68 & \bf 79.40  & \bf 82.21 & \bf 82.21 & \bf 82.21 \\
\hline
WNN     & word emb., suffix, capit. & 78.33 & 76.31 & 77.30  & 79.64 & 78.67 & 79.15 \\
\hline
WNN & word embeddings & 73.87 & 68.45 & 71.06 & 73.77 & 68.19 & 70.87 \\
\hline
CharNN  & char embeddings           & 53.86 & 51.40 & 52.60  & 61.13 & 59.03 & 60.06 \\

\hline
\end{tabular}
\end{center}
\end{table*}

In Table \ref{tab:esp:sota},
we compare CharWNN results with the ones of a state-of-the-art system for the SPA CoNLL-2002 Corpus.
This system was trained using AdaBoost and is described in \cite{carreras2002:conll}.
It employs decision trees as a base learner and uses handcrafted features as input.
Among others,
these features include gazetteers with people names and geographical location names.
The AdaBoost based system divide the NER task into two intermediate sub-tasks: NE identification and
NE classification. 
In the first sub-task, 
the system identifies NE candidates.
In the second sub-task,
the system classifies the identified candidates.
In Table \ref{tab:esp:sota},
we can see that even using only automatically learned features,
CharWNN achieves state-of-the-art results for the SPA CoNLL-2002.
This is an impressive result,
since NER is a challenging task to perform without the use of gazetteers.

\begin{table*}[ht!]
\begin{center}
\caption{Comparison with the state-of-the-art for the SPA CoNLL-2002 corpus.}
\label{tab:esp:sota}
\begin{tabular}{c|c|rrr}
\hline
\bf System & \bf Features         & \bf Prec. & \bf Rec.  & \bf F1 \\
\hline
  CharWNN & word embeddings, char embeddings & \bf 82.21 & \bf 82.21 & \bf 82.21 \\
\hline
          & words, ortographic, POS tags, trigger words,  & & & \\
 AdaBoost & bag-of-words, gazetteers, word suffixes, & 81.38 & 81.40 & 81.39  \\
          & word type patterns, entity length       & & & \\
\hline
\end{tabular}
\end{center}
\end{table*}

\subsection{Results for Portuguese NER}

In Table \ref{tab:por:nns},
we report the performance of different NNs for the HAREM I corpus.
The results in this table were computed using the CoNLL-2002 evaluation script.
We report results in two scenarios:
total and selective.
In the \emph{total} scenario, 
all ten categories are taken into account when scoring the systems. 
In the \emph{selective} scenario, 
only five chosen categories
(Person, Organization, Location, Date and Value) are taken into account.
We can see in Table \ref{tab:por:nns},
that CharWNN and WNN that uses two additional handcrafted features have similar results.
We think that by increasing the training data,
CharWNN has the potential to learn better character embeddings and outperform WNN, 
like happens in the SPA CoNLL-2002 corpus,
which is larger than the HAREM I corpus.
Again,
CharNN
and 
WNN that uses word embeddings only,
do not achieve results competitive with the results of the networks that jointly use word-level and character-level information.

\begin{table*}[ht!]
\begin{center}
\caption{Comparison of different NNs for the HAREM I corpus.}
\label{tab:por:nns}
\begin{tabular}{c|c|rrr|rrr}
\hline
\multirow{2}{*}{\bf NN} & \multirow{2}{*}{\bf Features} & \multicolumn{3}{c|}{\textbf{Total Scenario}} & \multicolumn{3}{c}{\textbf{Selective Scenario}} \\
\cline{3-8}
       & & \bf Prec. & \bf Rec. & \bf F1 & \bf Prec. & \bf Rec. & \bf F1\\
\hline
CharWNN & word emb., char emb.      & 67.16 & \bf 63.74 & 65.41 & 73.98 & \bf 68.68 & 71.23\\
\hline
WNN     & word emb., suffix, capit. & \bf 68.52 & 63.16 & \bf 65.73 & \bf 75.05 & 68.35 &  \bf 71.54\\
\hline
WNN     & word embeddings & 63.32 & 53.23 & 57.84 & 68.91 & 58.77 & 63.44 \\
\hline
CharNN  & char embeddings & 57.10 & 50.65 & 53.68 & 66.30 & 54.54 & 59.85 \\
\hline
\end{tabular}
\end{center}
\end{table*}

In order to compare CharWNN results with the one of the state-of-the-art system,
we report in tables \ref{tab:por:sota} and \ref{tab:por:byclass} the precision,
recall,
and F1 scores computed with the evaluation scripts from the HAREM I competition\footnote{http://www.linguateca.pt/primeiroHAREM/harem\_Ar\-quitectura.html} \cite{santosdiana07},
which uses a scoring strategy different from the CoNLL-2002 evaluation script.

In Table \ref{tab:por:sota},
we compare CharWNN results with the ones of ETL$_{CMT}$,
a state-of-the-art system for the HAREM I Corpus \cite{santos2012:book}.
ETL$_{CMT}$ is an ensemble method that uses Entropy Guided Transformation Learning (ETL) as the base learner.
The ETL$_{CMT}$ system uses handcrafted features like gazetteers and dictionaries as well as the output of other NLP tasks such as POS tagging and noun phrase (NP) chunking.
As we can see in Table \ref{tab:por:sota},
CharWNN outperforms the state-of-the-art system by a large margin in both total and selective scenarios,
which is an remarkable result for a system that uses automatically learned features only.

\begin{table*}[ht!]
\begin{center}
\caption{Comparison with the State-of-the-art for the HAREM I corpus.}
\label{tab:por:sota}
\begin{tabular}{c|c|rrr|rrr}
\hline
\multirow{2}{*}{\bf System} & \multirow{2}{*}{\bf Features} & \multicolumn{3}{c|}{\textbf{Total Scenario}} & \multicolumn{3}{c}{\textbf{Selective Scenario}} \\
\cline{3-8}
       & & \bf Prec. & \bf Rec. & \bf F1 & \bf Prec. & \bf Rec. & \bf F1\\
\hline
CharWNN       & word emb., char emb.        & 74.54 & \bf 68.53 & \bf 71.41 & \bf 78.38 & \bf 77.49 & \bf 77.93\\
\hline
              & words, POS tags, NP tags, & & & & & & \\
ETL$_{CMT}$     & capitalization, word length, & \bf 77.52 & 53.86 & 63.56 & 77.27 & 65.20 & 70.72 \\
              & dictionaries, gazetteers  & & & & & & \\
\hline
\end{tabular}
\end{center}
\end{table*}


In Table \ref{tab:por:byclass},
we compare CharWNN results by entity type with the ones of ETL$_{CMT}$.
These results were computed in the selective scenario.
CharWNN produces a much better recall than ETL$_{CMT}$ for the classes LOC,
PER and ORG.
For the ORG entity,
the improvement is of 21 points in the recall.
We believe that a large part of this boost in the recall is due to the unsupervised pre-training of word embeddings, 
which can leverage large amounts of unlabeled data to produce reliable word representations.

\begin{table*}[ht!]
\begin{center}
\caption{Results by entity type for the HAREM I corpus.}
\label{tab:por:byclass}
\begin{tabular}{c|rrr|rrr}
\hline
\multirow{2}{*}{\bf Entity} & \multicolumn{3}{c|}{\textbf{CharWNN}} & \multicolumn{3}{c}{\textbf{ETL$_{CMT}$}} \\
\cline{2-7}
       & \bf Prec. & \bf Rec. & \bf F1 & \bf Prec. & \bf Rec. & \bf F1\\
\hline
DATE & 90.27 & 81.32 & 85.56 & 88.29 & 82.21 & 85.14 \\
LOC & 76.91 & 78.55 & 77.72 & 76.18 & 68.16 & 71.95 \\
ORG & 70.65 & 71.56 & 71.10 & 65.34 & 50.29 & 56.84 \\
PER & 81.35 & 77.07 & 79.15 & 81.49 & 61.14 & 69.87 \\
VALUE & 78.08 & 74.99 & 76.51 & 77.72 & 70.13 & 73.73 \\
\hline
Overall & 78.38 & 77.49 & 77.93 & 77.27 & 65.20 & 70.72 \\
\hline
\end{tabular}
\end{center}
\end{table*}

\subsection{Impact of unsupervised pre-training of word embeddings}
In Table \ref{tab:esp:wordemb}
we assess the impact of unsupervised pre-training of word embeddings in CharWNN performance for both SPA CoNLL-2002 and HAREM I (selective).
The results were computed using the CoNLL-2002 evaluation script.
For both corpora,
CharWNN results are improved when using unsupervised pre-training.
The impact of unsupervised pre-training is larger for the HAREM I corpus (13.2 points in the F1) than for the SPA CoNLL-2002 (4.3 points in the F1).
We believe one of the main reasons of this difference in the impact is the training set size,
which is much smaller in the HAREM I corpus.

\begin{table*}[ht!]
\begin{center}
\caption{Impact of unsup. pre-training of word emb. in CharWNN performance.}
\label{tab:esp:wordemb}
\begin{tabular}{l|l|rrr}
\hline
Corpus & Pre-trained word emb.& \bf Precision & \bf Recall & \bf F1 \\
\hline
\multirow{2}{*}{SPA CoNLL-2002} & Yes & \bf 82.21 & \bf 82.21 & \bf 82.21  \\
                                & No  & 78.21 & 77.63 & 77.92 \\
\hline
\multirow{2}{*}{HAREM I}        & Yes & \bf 73.98 & \bf 68.68 & \bf 71.23 \\
                                & No  & 65.21 & 52.27 & 58.03 \\
\hline
\end{tabular}
\end{center}
\end{table*}


\section{Related Work} \label{sec:relatedwork}
Some recent work on deep learning for named entity recognition include Chen et al. \shortcite{chen:2010},
Collobert et al. \shortcite{collobert:jmlr2011}
and Passos et al. \shortcite{passos:2014}.

Chen et al. \shortcite{chen:2010} employ deep belief networks (DBN) to perform named entity categorization.
In their system, 
they assume that the boundaries of all the entity mentions were previously identified,
which makes their task easier than the one we tackle in this paper.
The input for their model is the character-level information of the entity to be classified.
They apply their system for a Chinese corpus and achieve state-of-the-art results for the NE categorization task.

Collobert et al. \shortcite{collobert:jmlr2011} propose a deep neural network which is equivalent to the WNN architecture described in Section \ref{model:setup}.
They achieve state-of-the-art results for English NER by adding a feature based on gazetteer information.

Passos et al. \shortcite{passos:2014} extend the Skip-Gram language model \cite{mikolov:icrl2013} to produce \emph{phrase embeddings} that are more suitable to be used in a linear-chain CRF to perform NER.
Their linear-chain CRF,
which also uses additional handcrafted features such as gazetteer based,
achieves state-of-the-art results on two English corpora:
CoNLL 2003 
and Ontonotes NER.

The main difference between our approach and the ones proposed in previous work is the use of neural character embeddings. This type of embedding allows us to achieve state-of-the-art results for the full task of identifying and classifying named entities using only features automatically learned. Additionally,
we perform experiments with two different languages,
while previous work focused in one language.
\section{Conclusions} \label{sec:conclusions}
In this work we approach language-independent NER using a DNN that employs word- and character-level embeddings to perform sequential classification.
We demonstrate that the same DNN which was successfully applied for POS tagging can also achieve state-of-the-art results for NER, 
using the same hyperparameters, 
and without any handcrafted features.
Moreover, 
we shade some light on the contribution of neural character embeddings for NER;
and define new state-of-the-art results for Portuguese and Spanish NER.




\bibliographystyle{acl}
\bibliography{ner_news2015}

\end{document}